\documentclass[11pt]{article}

\usepackage[final]{acl}

\usepackage{times}
\usepackage{latexsym}

\usepackage[T1]{fontenc}

\usepackage[utf8]{inputenc}

\usepackage{microtype}

\usepackage{inconsolata}

\usepackage{graphicx}

\usepackage{listings}

\title{GRACE: Grounded Adversarial Reasoning over Canadian Law}

\author{
  \textbf{Jiakang Xu\textsuperscript{1}\thanks{Equal contribution.}},
  \textbf{Wantong Huo\textsuperscript{1}\footnotemark[1]},
  \textbf{Udom Silparcha\textsuperscript{2}},
  \textbf{Jonathan H. Chan\textsuperscript{3}}
\\
  \textsuperscript{1}University of Toronto,
  \textsuperscript{2}UTOA Computing Analytics Co., Ltd.,
\\
  \textsuperscript{3}King Mongkut's University of Technology Thonburi
\\
  \small{
    \texttt{\{amyj.xu, vanessa.huo\}@mail.utoronto.ca},
    \texttt{udom@utoa-thai.com},
    \texttt{jonathan@sit.kmutt.ac.th}
  }
}

\begin{document}
\maketitle
\begin{abstract}

Large language models have shown strong performance across a range of legal tasks, but existing benchmarks rarely evaluate the ability to take and defend a legal position, reason under incomplete information, or synthesize multiple statutory provisions. This gap is particularly pronounced for Canadian law, which remains underrepresented in legal NLP. We introduce \textbf{GRACE} (\textbf{G}rounded \textbf{R}easoning \textbf{A}dversarial \textbf{C}anadian l\textbf{E}gal examples), a dataset of 1,915 question–reasoning–answer instances grounded in Canadian federal legislation. GRACE covers three reasoning modes: adversarial advocacy, uncertainty, and applied reasoning. We develop a pipeline that partitions raw statutory text, generates scenario-based questions and reasoning, and filters examples through model-free citation verification and LLM-based quality auditing. As a proof of concept, we fine-tune \textbf{CLeAR}-4B (\textbf{C}anadian \textbf{Le}gal \textbf{A}dversarial \textbf{R}easoning), a lightweight model for grounded legal reasoning, and evaluate it against the unmodified Qwen3-4B base model in open- and closed-book settings. CLeAR-4B substantially improves agreement with teacher outputs and statutory citation behavior when the relevant act text is provided, while its grounding degrades sharply when the statute is withheld. These results suggest that GRACE can support the development of lightweight legal models that reason more effectively from supplied statutory text.\footnote{Data and code: \url{https://anonymous.4open.science/r/GRACE-CLeAR}}
\end{abstract}

\section{Introduction}

Large language models (LLMs) have rapidly advanced across domains including healthcare, education, and law. In the legal domain, LLMs are pretrained or instruction fine-tuned on legal corpora and cases at scale to answer legal questions ~\citep{colombo2024saullm, niklaus-etal-2025-lawinstruct}. Others use RAG-based systems~\citep{disc-lawllm, chatlaw}, while some distill knowledge from a teacher model into a smaller student model~\citep{li-etal-2026-legaldrill}. These systems are often trained and evaluated on benchmarks such as LegalBench~\citep{legalbench} and MMLU~\citep{hendrycks2020measuring}, which include tasks covering legal rule recall, entailment, classification, and multiple-choice question answering. While these benchmarks capture a range of legal reasoning and information-seeking abilities, they rarely focus on evaluating a model's ability to take and defend a position or reason through an adversarial exchange. 

On the other hand, multi-agent systems such as AgentsCourt and AgentCourt ~\citep{he-etal-2024-agentscourt, chen-etal-2025-agentcourt} simulate courtroom proceedings by assigning different roles to LLM-based agents and generating adversarial interactions. These approaches demonstrate the potential of adversarial interaction for legal reasoning, but focus on agent-level simulation and evolution rather than training a standalone legal model on a dataset designed for adversarial debate. Moreover, Canadian law remains underrepresented in legal LLM research, with much of the existing work focusing on U.S. and Chinese legal systems. 

To our knowledge, no existing Canadian legal dataset is specifically designed to train or evaluate the reasoning required for adversarial legal debate. Existing Canadian datasets focus on legal corpora, retrieval, or question answering~\citep{wallace2025introducing, zhao2026canlegalragbench}. We address this gap by building a pipeline for constructing grounded legal reasoning data from raw statutory text and applying it to Canadian federal law. We summarize our main contributions as follows:

\begin{itemize}
    \item \textbf{Constructing a question-reasoning-answer dataset.}
    We construct \textbf{GRACE} (\textbf{G}rounded \textbf{R}easoning \textbf{A}dversarial \textbf{C}anadian l\textbf{E}gal examples), a dataset covering three reasoning modes: \textsc{Adversarial}, \textsc{Uncertainty}, and \textsc{Applied Reasoning}. To our knowledge, GRACE is the first Canadian legal dataset specifically designed for adversarial legal reasoning.
    \item \textbf{A pipeline for constructing grounded legal reasoning datasets.}
    We develop a pipeline that turns raw statutory text into grounded legal reasoning examples. The pipeline first partitions statutes into self-contained provisions, which are then used to generate scenario-based question–reasoning–answer instances. Each generated example is checked against a verbatim quote from the source text before it is included in GRACE. The pipeline can also be applied to statutory text from other jurisdictions. 
    \item \textbf{A proof-of-concept fine-tuned model.}
   As a proof of concept, we fine-tune Qwen/Qwen3-4B on GRACE to create a lightweight model for adversarial legal reasoning. We evaluate it against the unmodified base model under both open-book and closed-book settings, with the latter withholding the statute to test whether the model can reason from the supplied text rather than rely on memorized legal knowledge.
\end{itemize}

\begin{figure*}[t]
  \centering
  \includegraphics[width=\textwidth]{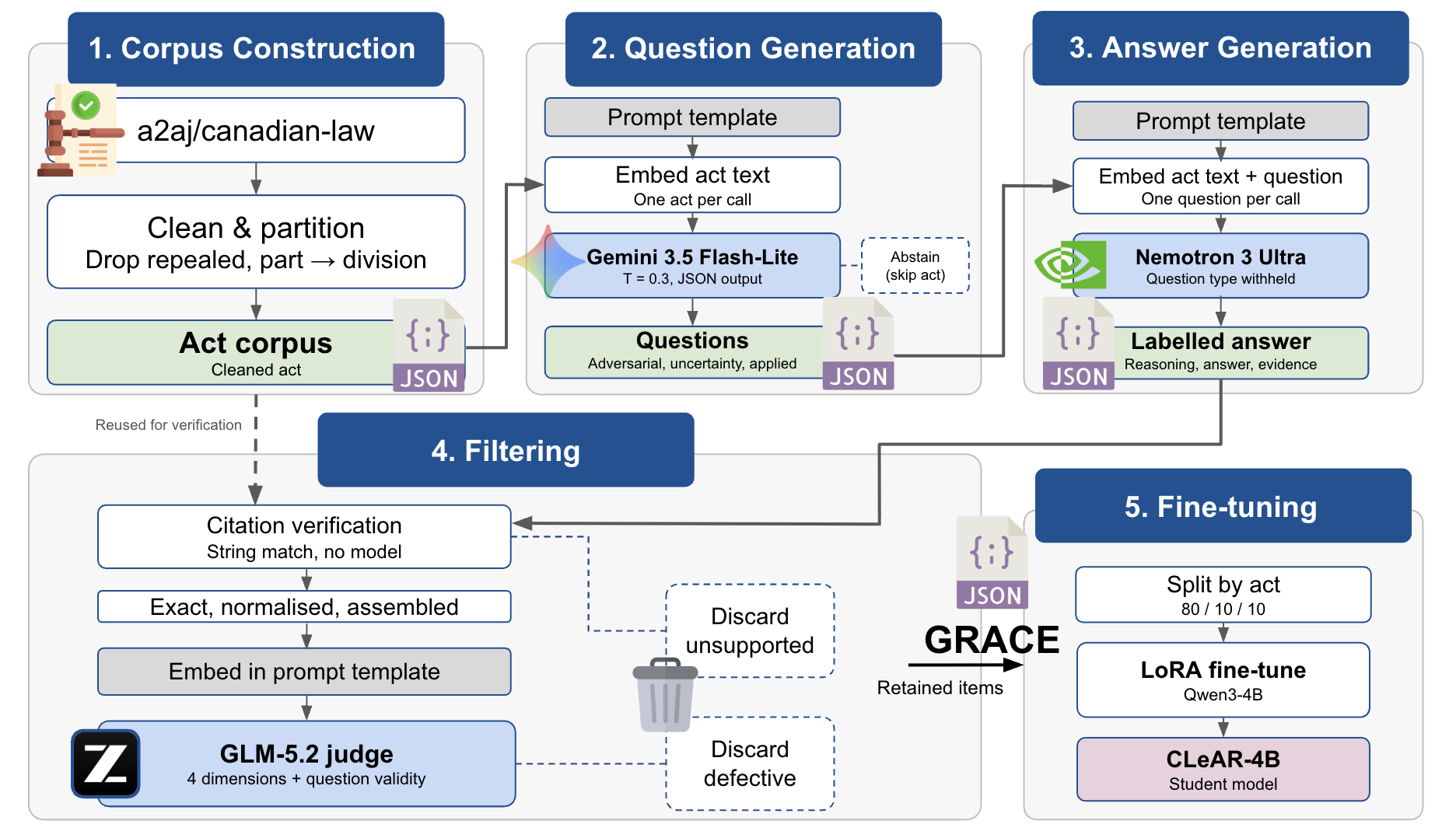}
  \caption{Pipeline diagram for constructing GRACE. Boxes in light green represent the content included in GRACE, boxes in light blue represent the LLMs used, and the box in pink represents the fine-tuned CLeAR-4B model.}
  \label{fig:pipeline}
\end{figure*}

\begin{figure*}[t]
  \centering
  \includegraphics[width=\textwidth]{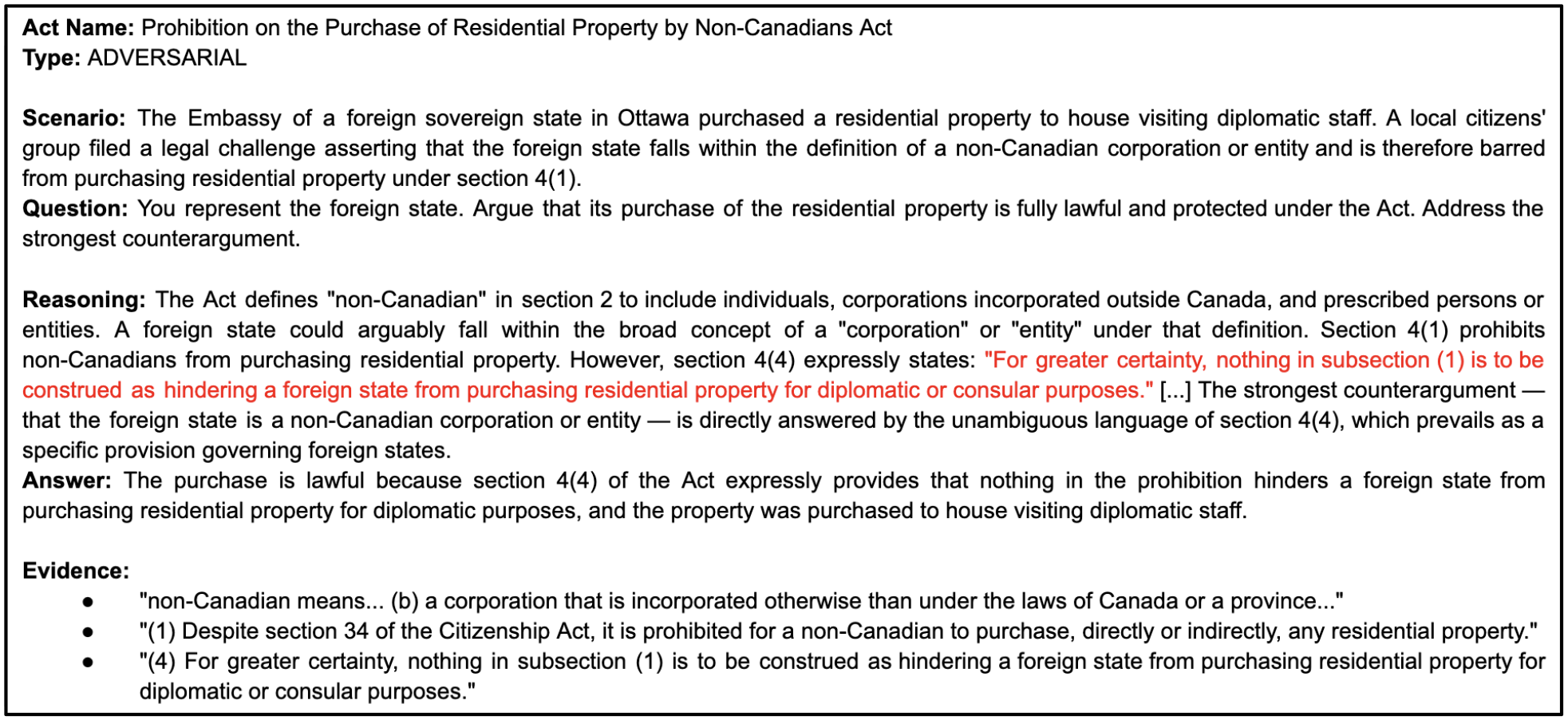}
  \caption{A representative ADVERSARIAL example from GRACE. The highlighted text is a verbatim quote from the source statute that resolves the counterargument. Question and reasoning are truncated for space. Full example shown in Appendix ~\ref{app:generated-example}.}
  \label{fig:sample_dataset}
\end{figure*}

\section{Related Work}
\subsection{Legal LLMs}
\citet{colombo2024saullm} introduced SaulLM, a domain-adapted language model further pretrained on large-scale legal corpora, improving performance on a range of legal understanding and generation tasks. \citet{niklaus-etal-2025-lawinstruct} explore instruction tuning approaches, where general-purpose language models are fine-tuned on curated legal instruction datasets to improve their ability to follow legal queries and perform structured reasoning. Other approaches incorporate external knowledge through retrieval. \citet{disc-lawllm} integrates document-level retrieval to ground model outputs in relevant legal texts, while \citet{chatlaw} combines retrieval with conversational interfaces to support interactive legal question answering grounded in case law and statutes. Finally, \citet{li-etal-2026-legaldrill} applies knowledge distillation to transfer capabilities from larger legal language models into smaller, more efficient models. 

\subsection{Agentic Court Simulation}

\citet{he-etal-2024-agentscourt} introduced AgentsCourt, a multi-agent framework that follows a courtroom process involving opening statements, debate, legal-resource retrieval, and judgment. The framework also introduces SimuCourt ~\citep{zhang2026chinese}, a benchmark built from Chinese judgment documents. More recently, \citet{chen-etal-2025-agentcourt} proposed AgentCourt, which uses adversarial interactions between lawyer agents to simulate court proceedings and accumulates experience from simulated cases through an evolving knowledge base. These approaches are closely related to our goal of modeling adversarial legal reasoning, but they approach the problem primarily through multi-agent simulation. In contrast, we focus on constructing a grounded dataset that can be used to train and evaluate a standalone model. 

\subsection{Legal Corpora and Benchmarks}

LegalBench ~\citep{legalbench} is a benchmark centered heavily on the United States legal system, covering a range of legal tasks: rule recall, entailment, classification, and other forms of legal reasoning.  JudiFair ~\citep{hu2026llms} is a dataset built on Chinese statutes designed to evaluate the systematic fairness of LLMs, specifically, inconsistency, bias, and imbalanced inaccuracy. Canadian legal NLP has received little attention than work focused on U.S. and Chinese legal systems. \citet{zhao2026canlegalragbench} introduced CanLegalRAGBench, a Canadian legal QA benchmark built from realistic queries and expert-annotated answers grounded in Canadian case law. However, no existing Canadian legal dataset is designed to train or evaluate a model on the process of taking and defending a legal position, a central part of adversarial legal debate.

\section{Methodology}

\subsection{Corpus Construction}
We construct our question, reasoning, and answer dataset from the public Canadian legislation dataset a2aj/canadian-laws ~\citep{wallace2025introducing}, which contains 22,785 statutes and regulations across federal, provincial and territorial jurisdictions. Due to computational constraints, we restrict our scope to federal statutes and regulations (25\% of the full dataset).

The acts' length distribution is heavily right-skewed. Manual inspection of the shortest documents shows that many are repealed acts or lacked sufficient information for downstream question generation, including placeholder documents whose entire body is a sentinel such as [Repealed, SOR/2008-90, s. 1] or title-only instruments. We therefore remove fully repealed samples, and samples with fewer than 255 characters.

For acts exceeding the context window of gemini-3.5-flash-lite, we partition the act at Part headings. Where a Part remains oversized, we further partition at Division headings. This introduces cross-reference limitations, as a partition may depend on definitions or provisions in another partition. We therefore allow the question generator to decline excerpts whose operative provisions depend too heavily on unavailable material.

\subsection{Question Generation}
We construct a question generation prompt (Appendix A) instructing gemini-3.5-flash-lite (temperature 0.3, JSON output enforced, and all tool use disabled to prevent external retrieval) to generate three question types: adversarial, uncertainty, and applied. Each processed act is embedded individually, with one act per call. We chose gemini-3.5-flash-lite for its strong performance on the Artificial Analysis Legal Index, particularly in non-hallucination and long-context reasoning ~\citep{artificialanalysis_capability_indices}.

Adversarial questions present a scenario based on the act and require a specific position to be defended, requiring the answering model to construct the best available case. These support downstream SLM debate, where fine-tuned small models take committed adversarial stances. Uncertainty questions present scenarios that deliberately omit information crucial to determining the answer, training the SLM to avoid unwarranted confidence. Applied questions require synthesizing multiple provisions or parts of an act to derive a single correct answer.

Each question includes its grounding: the output lists the atomic claims it relies on and verbatim citations from the act used to establish the scenario or question, enabling downstream verification without a model in the loop (§3.4).

We assign question counts based on act length. Adversarial questions form the majority to support the downstream debating task. Because these counts are static across varying text lengths, the prompt treats them as maxima rather than targets: Gemini may generate fewer questions when the act cannot support distinct, provided it records the reason. It may also decline an act entirely using one of five labelled reasons: insufficient text, external dependence, repealed or spent, amending-only, or corrupted text. Additional question-generation constraints are given in Appendix ~\ref{app:question-generation-prompt}.

\subsection{Answer Generation}
We construct an answer generation prompt (Appendix ~\ref{app:answer-generation-prompt}) for nvidia/nemotron-3-ultra-550b-a55b. We chose Nemotron mainly for its high non-hallucination score on the Artificial Analysis Legal Index and relatively strong reasoning performance ~\citep{artificialanalysis_capability_indices}, to ensure reliable and logically sound answer generation. Using a different model family reduces self-preference bias by ensuring that the answering model does not evaluate questions it generated itself. Nemotron receives the act, scenario, and question in an independent call for each item, preventing cross-item reasoning. It is instructed to ground answers in the act text and quote verbatim any provision used in its reasoning or answer. The question type is not disclosed: whether a scenario is under-determined must be detected by the model, making sufficiency a measurable calibration signal (§3.5).

\subsection{Citation Verification}
The integrity of the answers is central to dataset quality. We therefore first apply model-free filtering to verify that quoted citations occur in the source act. Manual inspection found that Nemotron often concatenates non-contiguous but individually valid excerpts, typically joining a provision’s opening stem with an applicable subparagraph while omitting intervening text. A flat substring check would therefore misclassify these legally valid selective quotations as fabricated. We therefore use tiered matching:

\begin{itemize}
    \item \textbf{exact} --- byte-identical substring of the act;
    \item \textbf{normalised} --- identical after collapsing whitespace and markdown emphasis;
    \item \textbf{cleaned} --- additionally removes presentation elements such as quotation marks, section notes, and terminal punctuation;
    \item \textbf{assembled} --- non-contiguous fragments whose individual components all occur verbatim in the act;
    \item \textbf{partial} --- at least half of the fragments are found;
    \item \textbf{unsupported} --- none of the above.
\end{itemize}

Generator and answerer citations are verified independently so failures can be attributed to the stage that produced them. We remove questions containing citations in the partial or unsupported tiers.

\subsection{LLM-as-Judge Filtering}
String matching verifies that a quotation is real but not whether the reasoning based on it is sound. We therefore use z-ai/glm-5.2 as an LLM judge from a third model family. GLM-5.2 ranks 10th on the Artificial Analysis Legal Index for legal reasoning and outperforms Nemotron ~\citep{artificialanalysis_capability_indices}, making it well suited as a stronger judge for evaluating Nemotron’s answers. The judge scores surviving answers against the full act on four dimensions: grounding (legal propositions are traceable to the act), soundness (reasoning supports the conclusion), task fit (the appropriate standard for adversarial, uncertainty, or applied questions), and completeness (all governing provisions are engaged). The judge must independently work through the statute—and, for applied questions, determine its own outcome—before reading the answer's conclusion, and must identify a specific defect for any sub-maximal score.

\begin{table*}[t]
  \centering
  \begin{tabular}{lcccc}
    \hline
    \textbf{System} & \textbf{ROUGE-1} & \textbf{ROUGE-2} & \textbf{ROUGE-L} & \textbf{BERTScore F1} \\
    \hline
    CLeAR-4B (Open)     & \textbf{0.6525} & \textbf{0.3973} & \textbf{0.4220} & \textbf{0.8992} \\
    CLeAR-4B (Closed)   & 0.5208 & 0.2391 & 0.2984 & 0.8688 \\
    Qwen3-4B (Open)   & 0.1999 & 0.1079 & 0.1501 & 0.8580 \\
    Qwen3-4B (Closed) & 0.1655 & 0.0700 & 0.1250 & 0.8329 \\
    \hline
  \end{tabular}
  \caption{Natural language generation similarity evaluation results across our fine-tuned (CLeAR-4B) model and base (Qwen3-4B) under open- and closed-book settings.}
  \label{tab:automatic-evaluation}
\end{table*}
\begin{table*}[t]
  \centering
  \begin{tabular}{lccccc}
    \hline
    \textbf{System} &
    \textbf{Quotes} &
    \shortstack{\textbf{Quotes}\\\textbf{per answer}} &
    \shortstack{\textbf{Verbatim}\\\textbf{rate}} &
    \shortstack{\textbf{Grounded}\\\textbf{rate}} &
    \shortstack{\textbf{Unsupported}\\\textbf{rate}} \\
    \hline
    CLeAR-4B (Open)   & 344 & 1.8 & \textbf{68.9\%} & \textbf{80.2\%} & \textbf{19.5\%} \\
    CLeAR-4B (Closed) & 273 & 1.4 & 15.8\% & 16.8\% & 83.2\% \\
    Qwen3-4B (Open)   & 57  & 0.3 & 66.7\% & 78.9\% & 21.1\% \\
    Qwen3-4B (Closed) & 184 & 0.9 & 3.3\%  & 3.3\%  & 96.7\% \\
    \hline
  \end{tabular}
  \caption{Grounding and citation statistics across fine-tuned (CLeAR-4B) and Qwen3-4B models under open- and closed-book settings.}
  \label{tab:grounding-evaluation}
\end{table*}

\begin{table*}[t]
  \centering
  \begin{tabular}{lcccc}
    \hline
    \textbf{System} & \textbf{Section recall (n)} & \textbf{Section recall} & \textbf{Section precision} \\
    \hline
    CLeAR-4B (Open)   & 193 & 78.1\% & 85.1\% \\
    CLeAR-4B (Closed) & 193 & 16.5\% & 44.6\% \\
    Qwen3-4B (Open)   & 193 & 57.9\% & 88.9\% \\
    Qwen3-4B (Closed) & 193 & 9.8\%  & 74.3\% \\
    \hline
  \end{tabular}
  \caption{Section-level recall and precision across fine-tuned (CLeAR-4B) and Qwen3-4B models under open- and closed-book settings.}
  \label{tab:section-retrieval}
\end{table*}

The judge also flags defective questions, including contradictory scenarios, questions unanswerable from the supplied text, uncertainty questions whose facts actually determine the outcome, and questions that leak the expected conclusion.

Judging covers an audited sample rather than the full dataset. We remove items when any dimension scores 2 or below. Unaudited items are retained because lack of auditing does not imply defectiveness. Since judging is sampled, the resulting defect rates characterize dataset quality rather than guarantee the quality of every retained item.

\subsection{Fine-tuning}
The two filtering stages produce question–reasoning–answer pairs, which we split by act rather than question in an 80/10/10 ratio. Splitting by act prevents questions sharing statutory provisions and defined terms from appearing across splits, which would otherwise inflate held-out performance through near-duplicates.

We fine-tune Qwen3-4B with LoRA (rank 16, $\alpha=32$, dropout 0.05). We select Qwen3-4B: openly available, independent of the generator and answerer, deployable on a single GPU, and with a 32K context window comfortably exceeding our statutory excerpts. Training is open-book: each example provides the act text, scenario, and question, with the target consisting of the teacher's REASONING followed by its ANSWER. Including the act mirrors the teacher's answering conditions and encourages grounding in provided text rather than memorization. Training on reasoning rather than answers alone is intended to distill the reasoning model's derivation process.

Because no established benchmark exists for Canadian legal reasoning, we evaluate against the teacher's outputs on held-out acts under four conditions: the fine-tuned adapter and the unmodified base model, each open-book and closed-book. The closed-book condition withholds the statute; since training was exclusively open-book, it separates reasoning grounded in the supplied text from statutes memorised during pretraining. All four run on identical questions with identical decoding. We report ROUGE ~\citep{lin-2004-rouge} and BERTScore~\citep{zhang2019bertscore} against the teacher's reasoning and answer, treating them as proxies for surface overlap rather than legal correctness, and additionally apply the tiered citation matching of §3.4 to the student's own quotations to test whether fine-tuning transferred the teacher's grounding discipline or only its style.

\section{Dataset}
GRACE consists of 1,915 grounded legal-reasoning examples synthesized from Canadian federal statutes, drawn from 272 distinct acts and regulations in a
corpus chunked from the \texttt{a2aj/canadian-laws} dataset~\citep{wallace2025introducing}. We split these by source statute into 1,530 training, 190 validation, and
195 test examples, so that no statute's questions cross splits. Each example pairs a fictional scenario and question with a real statutory excerpt, a reasoning chain, and an answer. Examples target one of three reasoning modes: adversarial advocacy (914 examples), applied synthesis of multiple provisions (667 examples), or reasoning under uncertainty (334 examples). Every example is grounded in the source statute through 1--10 verbatim quotations, with an average of 2.4 quotations per example. Questions were generated by Gemini-3.5-Flash-Lite and answered by Nemotron-3 Ultra-550B. Examples flagged by an LLM-judge (GLM-5.2 judge) as defective during quality auditing were removed before inclusion. Example samples are shown in Figure~\ref{fig:sample_dataset}.

We designed GRACE around three goals:
\begin{itemize}
    \item \textbf{Grounded reasoning.}
    Questions and answers are tied directly to the underlying statutory text, allowing the reasoning to be checked against the law rather than relying on unsupported or fake legal knowledge.
    \item \textbf{Multiple reasoning modes.}
    GRACE goes beyond single-turn legal question answering by requiring models to argue for an assigned position, recognize when the available facts are insufficient for a conclusion, and synthesize multiple statutory provisions to reach an outcome.
    \item \textbf{Quality-controlled examples.}
    The construction process combines automatic citation verification with LLM-based auditing to identify unsupported reasoning, defective questions, and mismatches between questions and their intended reasoning mode.
\end{itemize}

\section{Results}
\label{sec:results}
We evaluate two models on the held-out test split (195 examples): the base model, which is Qwen3-4B, and our fine-tuned \textbf{CLeAR}-4B (\textbf{C}anadian \textbf{Le}gal \textbf{A}dversarial \textbf{R}easoning) model from the base model. Each system is evaluated under two conditions: \textit{open-book}, in which the source act text is included in the prompt, and \textit{closed-book}, in which it is withheld and the model must rely on whatever legal knowledge it retains internally.

\subsection{Natural Language Generation Similarity}
We evaluate Natural Language Generation (NLG) similarity against the
reference answers in GRACE, generated by nvidia/nemotron-3-ultra-550b-a55b, as shown in Table~\ref{tab:automatic-evaluation}. ROUGE-1 and ROUGE-2 measure the overlap of individual words and word pairs, respectively, while ROUGE-L measures the longest common subsequence between the generated and reference answers~\citep{lin-2004-rouge}. BERTScore F1 uses contextual embeddings to measure semantic similarity rather than relying on exact word overlap
~\citep{zhang2019bertscore}. In the open-book condition, CLeAR-4B scores higher than the untrained Qwen3-4B baseline on all four metrics: ROUGE-1 (0.6525 vs.\ 0.1999), ROUGE-2 (0.3973 vs.\ 0.1079), ROUGE-L (0.4220 vs.\ 0.1501), and BERTScore F1 (0.8992 vs.\ 0.8580). CLeAR-4B also outperforms Qwen3-4B on all four metrics in the closed-book condition. Both models score lower in the closed-book condition than in the open-book condition across all metrics.

\subsection{Citation Faithfulness}
We extract all quoted spans from its output and compare them against the source statutory text. We report the total number of quotes, the average number of quotes per answer, and the proportion of quotes in three categories: \textit{verbatim} (exact or whitespace-normalized matches), \textit{grounded} (verbatim matches plus reformatted or non-contiguous spans that can be traced), and \textit{unsupported} (quotes that cannot be traced). As shown in Table~\ref{tab:grounding-evaluation}, CLeAR-4B produces more quotes per answer than Qwen3-4B in the open-book setting (1.8 vs.\ 0.3), while achieving similar verbatim rates and grounded rates. In the closed-book setting, CLeAR-4B again produces more quotes per answer (1.4 vs.\ 0.9) and has higher verbatim and grounded rates. Both models produce substantially more unsupported quotes when the statute is withheld. 

\subsection{Section-Citation Overlap}
As shown in Table~\ref{tab:section-retrieval}, section recall measures how many of the statutory sections used in the reference answer were also cited by the system. Section precision measures how many of the sections cited by the system were also used in the reference answer. The recall denominator ($n = 193$) is the same across all four conditions because it is based on the reference answers, not the system being evaluated. In contrast, the precision denominator varies because it depends on how many sections each system cites.

\section{Discussion}
The results show a clear benefit from fine-tuning when the relevant statutory text is provided to the model. In the open-book setting, CLeAR-4B substantially outperforms the untrained Qwen3-4B baseline on all NLG similarity metrics, with ROUGE-L increasing from 0.1501 to 0.4220. The improvement is not limited to surface similarity: CLeAR-4B also produces substantially more statutory citations per answer and identifies relevant statutory sections more effectively. These results suggest that the model has learned to use the provided legal text in a way that is more consistent with the reference answers, rather than simply producing more legal-sounding text.

However, the closed-book results show the limitation: Since CLeAR-4B is fine-tuned with act text given, with the act text withheld, both models perform substantially worse, but their failure patterns differ. The untrained Qwen3-4B baseline frequently produces incomplete or malformed responses, resulting in very few valid grounded citations. CLeAR-4B, in contrast, continues to produce fluent legal-style responses and attempts to cite statutory provisions, but a much larger fraction of these citations cannot be verified against the source text. In particular, CLeAR-4B's unsupported citation rate increases from 19.5\% in the open-book setting to 83.2\% in the closed-book setting. The citation analysis provides further evidence of this pattern. In the closed-book setting, CLeAR-4B produces 1.4 quoted spans per answer on average, compared with 0.9 for Qwen3-4B, but only 16.8\% of CLeAR-4B's citations can be grounded in the source text. This difference is important because fluent output can make unsupported legal answer harder to identify than visibly incomplete output.

The section-level analysis shows a similar dependence on access to the source text. In the open-book setting, CLeAR-4B recovers a substantially larger portion of the statutory sections used by the reference answers. When the act text is withheld, section recall drops sharply. Fine-tuning alone does not appear to provide reliable recall of specific statutory provisions.

Taken together, these results support an open-book or retrieval-grounded deployment setting for CLeAR-4B. The model benefits substantially from being given the relevant statute, while closed-book use introduces a significant risk of unsupported statutory citations. 

\section{Future Works}

We view GRACE and fine-tuned CleAR-4B as a foundation for a multilayer debate and judging system. We envision extending this work by deploying multiple fine-tuned small language models (SLMs), each specializing in different areas of law and trained from different teacher models, providing complementary legal knowledge and analytical perspectives. These SLMs would independently analyze a legal case and present their arguments to multiple judges, which would evaluate the analyses and provide feedback for subsequent rounds of refinement. Through iterative debate and evaluation, the system would systematically work toward a well-supported conclusion.

This setup is intended to simulate a room of lawyers analyzing a case: individuals with different areas of expertise present and challenge competing analyses before collectively reaching a conclusion. While this work focuses on fine-tuning SLMs for legal reasoning, the underlying debate framework could extend to other domains where specialized models must evaluate competing perspectives and collaboratively reach a thoroughly considered conclusion.

\section{Conclusion}

We presented GRACE, a grounded dataset for adversarial legal reasoning over Canadian federal law, together with a pipeline for constructing and quality-controlling reasoning examples from raw statutory text. GRACE covers adversarial advocacy, uncertainty, and applied reasoning, with examples grounded through verbatim statutory citations and filtered through citation verification and LLM-based auditing.

As a proof of concept, we fine-tuned CLeAR-4B on GRACE and compared it with the Qwen3-4B base model under open- and closed-book settings. The fine-tuned model shows substantially stronger agreement with the teacher outputs and more effective use of relevant statutory provisions when the source text is provided. However, its performance and citation faithfulness decline substantially when the statute is withheld, indicating that fine-tuning on open-book examples does not provide reliable closed-book recall of specific legal provisions.

These results support GRACE as a foundation for developing lightweight, retrieval-grounded models for adversarial legal reasoning. Future work can extend the dataset to additional Canadian jurisdictions and expert-validated examples, while exploring multi-model debate and judging systems that allow specialized models to challenge and verify competing legal arguments.

\section*{Limitations}
GRACE has several limitations: First, due to computational constraints, we restrict the source corpus to Canadian federal legislation rather than using the full a2aj/canadian-laws dataset~\citep{wallace2025introducing}. As a result, the dataset does not cover provincial or territorial legislation, and models trained on GRACE may not generalize to those jurisdictions. Second, we use a single model for question and answer generation rather than generating multiple candidate answers with different models. Using several models could provide greater diversity in reasoning styles and reduce dependence on the behavior of any single generator. Similarly, we use the available generation model under practical cost constraints; stronger models could potentially produce higher-quality questions and reasoning examples. Finally, the dataset has not undergone full expert validation. We contacted a Canadian legal expert to manually review a subset of the generated examples as a quality check. However, due to limited expert availability, this review was not completed before the submission deadline. We therefore do not claim that the entire dataset has been manually verified by legal experts. 

\section*{Ethics}
GRACE is intended for research on grounded legal reasoning and is not designed to provide legal advice. Since errors in legal reasoning can have consequential effects, model outputs should not be treated as authoritative legal conclusions without verification against the underlying legislation and qualified legal expertise. Our dataset is constructed from publicly available Canadian federal legislation and does not intentionally include private or confidential personal information. However, scenarios and reasoning are generated by large language models and may contain errors or biases in the source material or generating models. We therefore apply citation verification and LLM-based quality auditing, but these procedures do not constitute legal expert validation of every example. The closed-book experiments further show that the fine-tuned model can produce fluent but unsupported citations, highlighting the risk of relying on the model without access to source text. We recommend using CLeAR-4B in retrieval- or open-book settings where its legal claims can be checked against the legislation.

\section*{Acknowledgements}
We gratefully acknowledge the support of the Mitacs Globalink Research Award and the Innovative Cognitive Computing Research Center for this work.


\bibliography{custom}

\appendix
\section{Question Generation Prompt}
\label{app:question-generation-prompt}

We construct a question generation prompt to generate grounded legal
reasoning questions from individual legislative acts. The full prompt is
shown below.

\begin{lstlisting}
You are a Canadian legal question-writer. Given the full text of a Canadian
legislative act, generate legal reasoning questions grounded strictly in it.

INPUT
Act name: {{ACT_NAME}}
Jurisdiction: {{JURISDICTION}}
Act text:
"""
{{ACT_TEXT}}
"""

RULES
- Answerable using ONLY the provided act text plus legal reasoning, no
  invented facts or definitions.
- Scenarios: realistic, specific (named parties, dates, facts), within the
  act's actual subject matter.
- Vary difficulty and which parts of the act are used; don't cluster on the
  same passage.
- Plain, professional legal English (law student / junior associate level),
  unless quoting statutes directly.
- `claims` lists each atomic claim the item relies on - a claim may be
  explicit (a fact or rule stated directly in the act) or interpretive
  (a reading that requires connecting two or more passages, e.g. a
  definition plus an operative clause).
- `evidence_of_claims` lists the exact quotes those claims come from - every
  passage of ACT_TEXT relied on anywhere in the item, whether explicit or
  implicit, and whether it's used in `claims`, the `scenario`, or the
  `question` itself. Each entry must be copied character-for-character from
  ACT_TEXT.
- A section number, a heading, or a Part title on its own is NOT a valid
  `evidence_of_claims` entry. Quote the words of the provision, not the
  marker that identifies it. Wrong: "7". Wrong:
  "PART II -- Description of the Outer Limits of Lands". Right:
  "No owner or occupier of any land or land under water to which these
  Regulations apply shall permit such land or any part thereof to be used
  for the disposal or accumulation thereon of any waste, material or
  substance edible by or attractive to birds."
- {{NUM_ADVERSARIAL}}, {{NUM_UNCERTAINTY}}, {{NUM_APPLIED}} are maximums,
  not required counts. Generate fewer for a given type only if the act
  does not support that many distinct, non-redundant items of that type
  (no reused scenario, no near-duplicate quote). Do not pad with weak or
  repetitive questions to hit the count.
- Output ONLY valid JSON per the schema below.

SKIP CONDITIONS
- Do NOT generate questions if any of the following apply. Instead output
  only:
  {"skip_reason": "<one of the labels below>"}

1. insufficient_text: ACT_TEXT is too short or purely administrative
   (title, definitions, commencement only) to support substantive questions.
2. external_dependence: operative provisions depend so heavily on
   instruments not provided (regulations, schedules, other statutes) that
   no question type could be answered from ACT_TEXT alone. Mere external
   references are fine - that is valid UNCERTAINTY material.
3. repealed_or_spent: ACT_TEXT indicates the act is repealed or expired.
4. amending_only: the act solely amends other statutes and cannot be
   understood without them.
5. corrupted_text: ACT_TEXT is garbled, truncated mid-provision, or not
   actually legislation.

- If none apply, you must generate the full set of questions.

QUESTION TYPES

1. PRIORITIZE GENERATING QUALITY QUESTIONS: ADVERSARIAL
   ({{NUM_ADVERSARIAL}}) - tests persuasive, position-committed argument.
   Give a scenario, assign a side (sometimes the stronger reading,
   sometimes the weaker one, so the model must build the best available
   case, not just state the obvious answer), require reliance on the act's
   text.

   Template:
   "[Scenario]. You represent [party]. Argue that [claim], relying on
   {{ACT_NAME}}. Address the strongest counterargument."

2. UNCERTAINTY ({{NUM_UNCERTAINTY}}) - tests calibration: whether the
   answering model NOTICES that it cannot reach a conclusion. Give a
   scenario that leaves out a fact the act needs to resolve it.

   THE QUESTION MUST NOT REVEAL THAT ANYTHING IS MISSING.

   Wrong:
   "What conclusion can be reached, what critical fact is missing, and why
   must a final determination be withheld?"

   Right:
   "Determine whether the Minister's direction is valid and enforceable
   against Arthur."

3. APPLIED REASONING ({{NUM_APPLIED}}) - tests reaching the single correct
   answer under complete facts, often synthesizing 2+ passages (rule +
   exception, definition + operative clause). Ask for the outcome plus the
   reasoning chain.

OUTPUT SCHEMA
{
  "act_name": "act name given in the prompt",
  "questions": [
    {
      "id": "ADV-01 / UNC-01 / APP-01",
      "claims": [
        "atomic claim the question relies on, e.g. 'notice must be given within 30 days'",
        "one claim per entry, each verifiable as true/false against the act text"
      ],
      "scenario": "<fictional facts only: named parties, dates, events. No task or question included.>",
      "question": "question based on the context, question only",
      "evidence_of_claims": [
        "the full operative wording of a provision, copied verbatim from ACT_TEXT - support for anything used in claims, scenario, or question above. Never a bare section number or heading."
      ]
    }
  ],
  "generation_summary": {
    "adversarial_generated": 0,
    "adversarial_requested": 0,
    "uncertainty_generated": 0,
    "uncertainty_requested": 0,
    "applied_generated": 0,
    "applied_requested": 0
  }
}

Include a "<type>_shortfall_reason" key in generation_summary
(e.g. "uncertainty_shortfall_reason") only for types where generated <
requested, explaining why no further distinct, non-redundant items could
be produced.

No two questions in the same type reuse a scenario or test the same narrow
part of the act identically.
\end{lstlisting}

\section{Answer Generation Prompt}
\label{app:answer-generation-prompt}
\begin{lstlisting}
You are a Canadian legal reasoning assistant. Answer the question below using only the provided act text.

INPUT
Act name: {{ACT_NAME}}
Jurisdiction: {{JURISDICTION}}
Act text:
"""
{{ACT_TEXT}}
"""

Scenario:
{{SCENARIO}}

Question:
{{QUESTION}}

RULES
- Ground every legal proposition in ACT_TEXT. Do not rely on outside legal knowledge, other statutes, or case law, even if you believe you know the answer.
- Advocacy must rest on the scenario's stated facts. Do not speculate about facts the scenario does not state ("the notice may have also required...") in order to strengthen your position. If the assigned position cannot succeed on the stated facts, say so in REASONING and make the best argument actually available.
- Some questions are fully resolvable on what you are given and some are not. Decide which case you are in before answering, and do not manufacture certainty the text and facts do not support. Where a required fact, definition, threshold, or external instrument is missing, name it specifically rather than gesturing at incompleteness.
- When assigned a side: your ANSWER argues that side as persuasively as the act allows, without misstating it. Your REASONING stays neutral: separate what the text clearly requires from what merely favors your side, give the counterargument its full weight, and if the text does not settle the question, say so plainly (e.g. "the text leaves X open; either reading is possible"). Do not dismiss a counterargument just because the act is silent on something - silence creates ambiguity, it does not win the point for you.
- Whenever you quote a passage of ACT_TEXT, quote it character-for-character - no paraphrase, no ellipses inside the quotation. Reserve quotations for REASONING and EVIDENCE; in ANSWER, cite provisions by section number only.
- Plain, professional legal English.

OUTPUT FORMAT
Write ordinary prose, but tag the following sections. Each label must appear in CAPITALS at the start of a new line, followed by a colon, in this order. Do not use these labels anywhere else in your response.

SUFFICIENCY: exactly one word - "sufficient" if ACT_TEXT plus the scenario's stated facts are enough to complete the task the question asks, or "insufficient" if a required fact, definition, threshold, or external instrument is missing. Where the question assigns you a side to argue, judge whether you can make the argument on the materials given, not whether the underlying dispute could be settled conclusively.
MISSING: what specifically would be needed to resolve the question. Write "none" when SUFFICIENCY is "sufficient".
REASONING: the legal analysis - which passages govern, how they apply to the facts, in the order a lawyer would work through them. Where a side has been assigned, this section is a neutral assessment of both readings and their relative strength; the one-sided presentation belongs in ANSWER only.
ANSWER: concise verdict; may restate the outcome and cite the controlling provisions, but must not repeat the Reasoning's analysis: no re-walking the facts element by element, no statutory quotations, no re-deriving conclusions already established. <=5 sentences.
EVIDENCE: the passages of ACT_TEXT your reasoning relies on. Put each quote on its own line beginning with "- ". Quote the wording exactly; where a passage runs across several lines or paragraphs in the act, join it onto one line with single spaces rather than paraphrasing or truncating it.
\end{lstlisting}

\section{A full example from GRACE}
\label{app:generated-example}
\begin{lstlisting}
Act Name: Prohibition on the Purchase of Residential Property by Non-Canadians Act
Type: ADVERSARIAL

Scenario: The Embassy of a foreign sovereign state in Ottawa purchased a residential property to house visiting diplomatic staff. A local citizens' group filed a legal challenge asserting that the foreign state falls within the definition of a non-Canadian corporation or entity and is therefore barred from purchasing residential property under section 4(1).
Question: You represent the foreign state. Argue that its purchase of the residential property is fully lawful and protected under the Act, relying on Prohibition on the Purchase of Residential Property by Non-Canadians Act. Address the strongest counterargument.

Reasoning: "The Act defines "non-Canadian" in section 2 to include individuals, corporations incorporated outside Canada, certain Canadian corporations controlled by non-Canadians, and prescribed persons or entities. A foreign state could arguably fall within the broad concept of a "corporation" or "entity" under that definition. Section 4(1) prohibits non-Canadians from purchasing residential property. However, section 4(4) expressly states: "For greater certainty, nothing in subsection (1) is to be construed as hindering a foreign state from purchasing residential property for diplomatic or consular purposes." This provision is a clear legislative directive that the prohibition does not apply to foreign states when the purchase is for diplomatic or consular purposes. The scenario states the Embassy of a foreign sovereign state purchased the property "to house visiting diplomatic staff," which constitutes a diplomatic purpose. Therefore, even if the foreign state were considered a non-Canadian, section 4(4) removes the application of the prohibition to this purchase. The strongest counterargument -- that the foreign state is a non-Canadian corporation or entity -- is directly answered by the unambiguous language of section 4(4), which prevails as a specific provision governing foreign states."
Answer: The purchase is lawful because section 4(4) of the Act expressly provides that nothing in the prohibition hinders a foreign state from purchasing residential property for diplomatic purposes, and the property was purchased to house visiting diplomatic staff. 

Evidence:
"non-Canadian means (a) an individual who is neither a Canadian citizen nor a person registered as an Indian under the Indian Act nor a permanent resident; (b) a corporation that is incorporated otherwise than under the laws of Canada or a province; (c) a corporation incorporated under the laws of Canada or a province whose shares are not listed on a stock exchange in Canada for which a designation under section 262 of the Income Tax Act is in effect and that is controlled by a person referred to in paragraph (a) or (b); and (d) a prescribed person or entity.", 
"(1) Despite section 34 of the Citizenship Act, it is prohibited for a non-Canadian to purchase, directly or indirectly, any residential property.", 
"(4) For greater certainty, nothing in subsection (1) is to be construed as hindering a foreign state from purchasing residential property for diplomatic or consular purposes."

\end{lstlisting}

\end{document}